%% file: main.tex
\pdfoutput=1
\PassOptionsToPackage{numbers,square,sort&compress}{natbib}
\PassOptionsToPackage{table,dvipsnames}{xcolor}
\documentclass[]{style/company_light} 

\usepackage[utf8]{inputenc}
\usepackage{hyperref}
\usepackage{url}
\usepackage{booktabs}

\usepackage{amsfonts}
\usepackage{amsmath}
\usepackage{nicefrac}
\usepackage{microtype}
\usepackage{graphicx}
\usepackage{float}
\usepackage{subcaption}
\usepackage{multirow}
\usepackage{enumerate}
\usepackage{enumitem}
\usepackage[most]{tcolorbox}
\usepackage{adjustbox}
\usepackage{array}
\usepackage{amssymb}
\usepackage{tabularx}
\usepackage{longtable}
\usepackage{listings}
\usepackage{needspace}

\newcounter{finding}
\newtcolorbox{findingbox}{
  enhanced,
  colback=brandprimary!12!white,
  colframe=brandprimary,
  coltext=black,
  boxrule=0.35pt,
  borderline west={3pt}{0pt}{brandaccent},
  sharp corners,
  boxsep=0pt,
  left=10pt, right=10pt, top=8pt, bottom=8pt,
  before skip=12pt, after skip=6pt,
  before upper={\refstepcounter{finding}},
  after={\par\nobreak\noindent}
}

\makeatletter
\def\fps@figure{htbp}
\def\fps@table{htbp}
\makeatother

\newcommand{\compactfloatlayout}{%
  \renewcommand{\topfraction}{0.92}%
  \renewcommand{\bottomfraction}{0.88}%
  \renewcommand{\textfraction}{0.08}%
  \renewcommand{\floatpagefraction}{0.85}%
  \setcounter{topnumber}{3}%
  \setcounter{bottomnumber}{2}%
  \setcounter{totalnumber}{4}%
  \setlength{\textfloatsep}{10pt plus 2pt minus 2pt}%
  \setlength{\intextsep}{10pt plus 2pt minus 2pt}%
  \setlength{\floatsep}{8pt plus 2pt minus 2pt}%
  \interlinepenalty=0\relax
  \clubpenalty=10000\relax
  \widowpenalty=10000\relax
  \raggedbottom
}

\NewDocumentCommand{\todo}
{ mO{} }{\textcolor{magenta}{\textsuperscript{\textit{TODO}}\textsf{\textbf{\small[#1]}}}}

\title{Frontier3DBench: Benchmarking Coding Agents for 3D Creation across Objects, Scenes, and Interactive Open Worlds}

\author[*]{Xinjie Lin}
\author[*]{Qingle Liu}
\author[*]{Wenyi Li}
\author{John Lam}
\author{Xiaoqiu Wang}
\author{Deyao Hong}
\author{Mingju Gao}
\author{Zheng Jiang}
\author{Yizhe Chi}
\author{Youjie Zheng}
\author[\ddagger]{Calvin Xiao}
\author[\ddagger]{Kaisen Yang}
\author[\ddagger]{Huan-ang Gao}

\affiliation{Navers Lab, Einsia.AI,\quad\quad Tsinghua University}
\contribution[*]{Equal Contribution}
\contribution[\ddagger]{Corresponding Authors}

\input{Sections/0_abstract}

\date{November 6, 2026}
\metadata[Homepage]{\url{https://lab.einsia.ai/frontier3dbench}}
\correspondence{\email{nana@einsia.ai}}

\begin{document}

\maketitle

\compactfloatlayout
\input{Sections/1_introduction}
\input{Sections/2_method}
\input{Sections/3_experiments}
\input{Sections/4_relatedworks}
\input{Sections/5_conclusion}
\par\FloatBarrier
\clearpage


\bibliographystyle{unsrtnat}
\begingroup
\renewcommand{\bibfont}{\fontsize{8.5}{10}\selectfont}
\setlength{\bibsep}{1pt plus 0.5pt}
\bibliography{ref}
\endgroup


\clearpage
\appendix
\input{Sections/appendix_score_weights}
\input{Sections/appendix_experiments}
\input{Sections/appendix_visual_results}

\end{document}

%% file: Sections/0_abstract.tex
\abstract{%
\setstretch{1.08}%
Coding agents combine visual understanding with program execution, offering a new route to automated 3D creation.
By translating images and instructions into editable programs, they bring reconstruction, generation, and editing into a shared workflow.
Yet existing benchmarks fragment evaluation across task scopes and protocols, leaving their capabilities across construction levels unclear.
We propose Frontier3DBench, a unified benchmark of 145 tasks across 9 categories spanning object, scene, and interactive open-world creation, with joint evaluation of appearance, structure, and behavior.
Experiments with 11 model--harness configurations reveal strengths that vary by level: GPT-6.1-sol leads the object and scene levels and achieves the highest overall score of 66.02/100, whereas GPT-6-astra leads interactive world construction.
However, all configurations score lower on reference reconstruction than on text-driven generation in object and scene tasks.
Producing plausible content thus remains easier than reproducing specified geometry, layouts, and visual details.
Moreover, generated worlds average 83.08/100 on local mechanisms but only 20.82 on motion clearance and 30.26 on stateful workflows: individual mechanisms often function, whereas complete activities that require coordinated actions and persistent state remain difficult.
These results identify reference fidelity and coordinated activity completion as central bottlenecks for coding agents in 3D creation.
\par
}

%% file: Sections/1_introduction.tex
\begin{figure}[!ht]
    \centering
    \includegraphics[width=\linewidth]{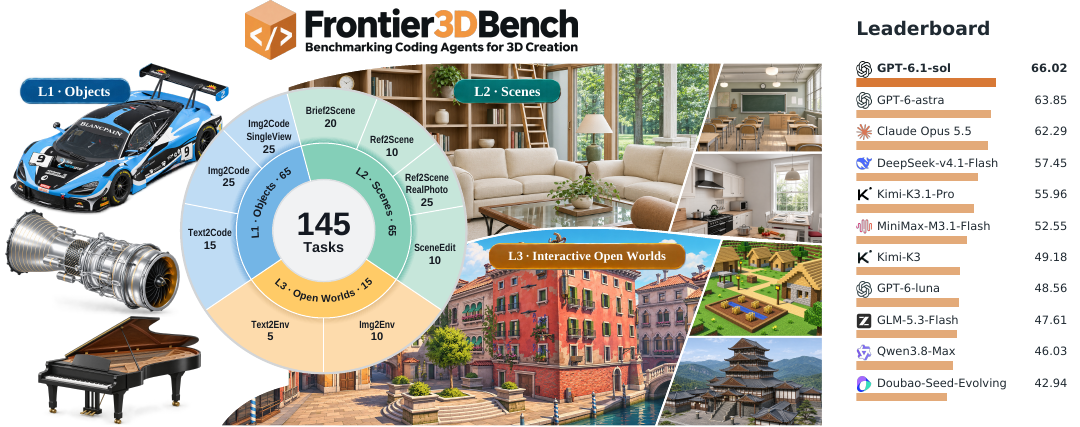}
    \caption{Frontier3DBench: 145 tasks in 9 categories spanning objects (L1), scenes (L2), and interactive open worlds (L3). Right: leaderboard of the 11 model--harness configurations (0--100, $\uparrow$).}
    \label{fig:teaser}
\end{figure}

\clearpage

\section{Introduction}
\label{sec:introduction}

Automating 3D creation has long been a central pursuit in computer vision, spanning reconstruction, generation, and editing~\cite{schonberger2016sfm,mildenhall2020nerf,kerbl2023gaussians,poole2023dreamfusion,lin2023magic3d,michel2022text2mesh,haque2023instructnerf2nerf}.
Yet these tasks demand more than recognizing visual content: they require translating observations and instructions into explicit geometry and spatial structure.
Coding agents offer a new route toward this goal because they combine strong visual understanding with the ability to write and execute programs that construct 3D content~\cite{viga2026,code2games2026}.
We argue that coding agents will replace traditional, task-specific computer vision pipelines for 3D creation.
The critical question is therefore: \textit{how far have coding agents come in 3D creation?}

Existing benchmarks, however, fragment this evaluation across different construction tasks and application settings.
For example, object and scene authoring benchmarks assess geometry, visual fidelity, construction, and editing~\cite{blenderworld2026,code4scene2026}, whereas executable-world and game benchmarks emphasize mechanical dependencies, interactive behavior, and gameplay completion~\cite{exeworld2026,code2games2026,cegame2026}.
Because their task scopes and evaluation protocols differ, these results are difficult to combine into a consistent account of an agent's capabilities across objects, scenes, and interactive worlds.
A single aggregate score can also obscure whether an agent fails to match a reference, organize a scene, or implement the required behavior.
Object fidelity, scene coherence, and world functionality impose distinct construction requirements.
A unified benchmark must therefore evaluate each level through dedicated tasks and complementary visual, structural, and behavioral checks.

Our key insight is that successful 3D creation demands faithful appearance, coherent structure, and functional behavior.
Accordingly, we organize evaluation around objects, scenes, and interactive open worlds, requiring executable, editable artifacts at every level.
Within this hierarchy, we assess object geometry and detail, scene layout and composition, and world-level appearance and activity completion through complementary visual, structural, and behavioral checks.
Rendered views measure appearance, native scene inspection measures structure, and runtime tests measure behavior.

Following these principles, we introduce \textbf{Frontier3DBench}, comprising 145 tasks across 9 categories: 65 object tasks, 65 scene tasks, and 15 interactive open-world tasks.
The suite covers generation, reconstruction, and scene editing in Blender, together with interactive world construction in Unreal Engine.
Evaluation of 11 model--harness configurations reveals three key findings.
First, model strengths vary by level: GPT-6.1-sol leads the object and scene levels and achieves the highest overall score of 66.02/100, whereas GPT-6-astra leads interactive world construction.
Second, every configuration scores lower on reference reconstruction than on text-driven generation in both object and scene tasks, highlighting the difficulty of matching specified geometry, layouts, and visual details.
Third, local mechanisms remain far stronger than coordinated activities: across all 165 L3 submissions (11 configurations $\times$ 15 tasks), the local-mechanism component of interaction correctness averages 83.08/100, compared with 20.82 for motion clearance and 30.26 for stateful workflows.
These findings point to reference fidelity and the coordination of moving parts, actions, and persistent state as the main open challenges in 3D creation.

Our contributions are threefold:
\begin{itemize}
    \item \textbf{A unified benchmark for coding agents in 3D creation.} Frontier3DBench comprises 145 tasks across 9 categories, bringing object, scene, and interactive-world construction into a shared suite covering generation, reconstruction, and scene editing. Dedicated tasks at each level reveal whether an agent's strengths extend from individual assets to complete environments.
    \item \textbf{Joint evaluation of appearance, structure, and behavior.} We assess executable Blender and Unreal Engine artifacts through visual judgments, structural measurements, and independent runtime tests, connecting content quality with action effects, state consistency, and activity completion.
    \item \textbf{An empirical account of current capabilities and bottlenecks.} Across 11 model--harness configurations, we identify strengths that vary by construction level and persistent gaps in reference fidelity and coordinated activity completion. We further analyze capability dimensions and API costs to characterize differences in performance and resource demands.
\end{itemize}

%% file: Sections/2_method.tex
\section{Frontier3DBench}
\label{sec:benchmark}

\subsection{Task Formulation}
\label{sec:task_formulation}

\begin{figure}[htbp]
  \centering
  \includegraphics[width=0.9\linewidth]{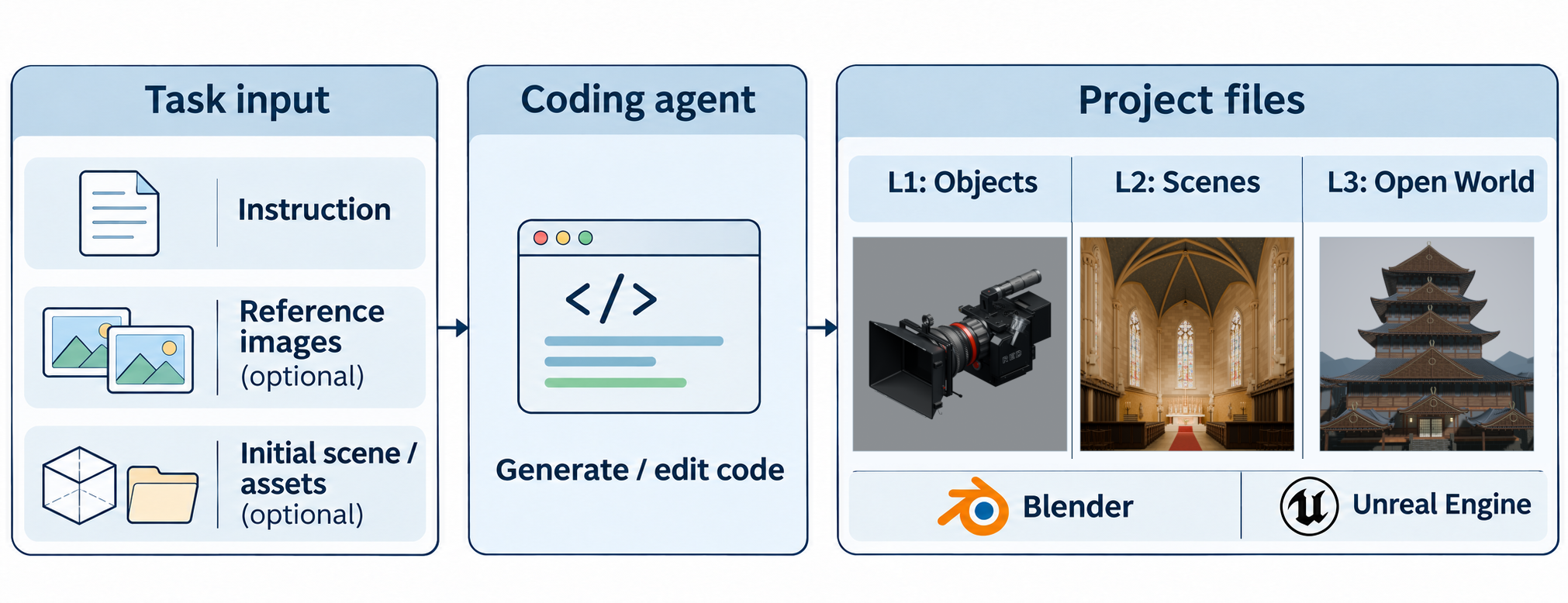}
  \caption{Task formulation. Agents receive instructions and optional references or initial assets, then produce code and assets for Blender objects and scenes (L1/L2) or Unreal Engine open worlds (L3).}
  \label{fig:task_formulation}
\end{figure}

Frontier3DBench formulates 3D creation as a mapping from a task specification to project files for the prescribed 3D engine (Figure~\ref{fig:task_formulation}).
These files encode the requested objects (L1), scenes (L2), or open worlds (L3), together with the code and assets needed to construct or run them.

\paragraph{Input.}
We represent the agent-visible input as $x=(d,\mathcal{I},a_0)$, where $d$ is a written task instruction, $\mathcal{I}$ is an optional set of reference images, and $a_0$ denotes any supplied starter code, initial scene, or reusable assets.
The instruction defines the required geometry, appearance, spatial relationships, and, for open worlds, connected regions and functional behavior.
Reference-conditioned tasks provide one or more images; text-conditioned tasks use $\mathcal{I}=\varnothing$.
Editing tasks additionally specify the initial scene, the requested modification, and the content to preserve.
For L3, the input includes an interaction contract specifying functional zones, entities, initial states, and available operations.
Source assets and evaluation annotations are used only for scoring and are not provided to the agent.

\paragraph{Output.}
Given $x$, a coding agent $\mathcal{A}$ produces a submission $y=\mathcal{A}(x;e)$ for the task's prescribed engine $e$:
\begin{equation}
e =
\begin{cases}
\text{Blender}, & \text{L1 objects and L2 scenes},\\
\text{Unreal Engine}, & \text{L3 open worlds}.
\end{cases}
\label{eq:task_engine}
\end{equation}
For L1 and L2, $y$ contains Blender project files for the resulting object or scene, executable construction or editing code, and task-required exports.
For L3, $y$ contains Unreal Engine project files for an open world that connects multiple regions into a coherent environment, including its assets and required interaction logic.
Executing the submission must realize the visual and structural requirements of the input; L3 must also support the specified actions and their persistent effects within a session.
An independent observer tests the required interactions and state persistence as the player moves through the generated world.

\subsection{Benchmark Construction}
\label{sec:benchmark_construction}

\subsubsection{Data Sources}

Object and scene tasks draw on publicly available 3D models (primarily \texttt{.glb}, with some \texttt{.blend} files), images from public datasets, and photographs captured by the authors.
Interactive open-world tasks use game screenshots and real-world photographs, including panoramic imagery, as visual references for interactive environments.
Text-conditioned tasks use text instructions without reference images.

\subsubsection{Benchmark Composition}
\label{sec:benchmark_composition}

\input{Figures/FigBenchmarkComposition}

\textbf{Overview.}
Frontier3DBench organizes 3D creation tasks by the unit being built and the behavior it must support.
The hierarchy progresses from individual objects and assemblies, through spatially organized scenes, to connected worlds with persistent interaction states.
Table~\ref{tab:benchmark_composition} summarizes the inputs and requirements of the 145 tasks across 9 categories and three levels.

\textbf{Level 1 (L1): Object Level.}
L1 concerns the construction of a single editable 3D asset, which can be an individual object or a mechanical assembly.
Task requirements address the asset's geometry, dimensions, appearance, and relationships among constituent parts.
The 65 tasks cover modeling from text (\textit{Text2Code}, 15), reconstruction from multiple reference views (\textit{Img2Code}, 25), and reconstruction from a single image (\textit{Img2Code-SingleView}, 25).

\textbf{Level 2 (L2): Scene Level.}
L2 concerns the construction or modification of a complete scene composed of multiple objects and their shared surroundings.
The objects must form a coherent spatial layout with consistent relative scales, while materials and lighting must support the specified appearance and camera composition.
The 65 tasks comprise scene generation from text instructions (\textit{Brief2Scene}, 20), reconstruction from reference images (\textit{Ref2Scene}, 10), reconstruction from real photographs (\textit{Ref2Scene-RealPhoto}, 25), and modification of an existing scene (\textit{SceneEdit}, 10).
Scene editing requires changing specified objects or scene properties while preserving the surrounding content.

\textbf{Level 3 (L3): Open-World Level.}
L3 concerns the construction of an explorable world that connects multiple functional regions within a shared environment.
The player must be able to move between regions and interact with objects and mechanisms whose state changes persist across action sequences within a session.
Complexity is defined by the required spatial organization and connected operations, allowing both dense indoor environments and larger outdoor worlds.
The 15 tasks are grouped by input modality: \textit{Text2Env} (5) uses a text instruction without reference images, while \textit{Img2Env} (10) pairs the instruction with one or more reference images.
Both categories require executable interactions that respect action prerequisites, so that changes produced by earlier actions determine which subsequent operations are available.

\subsubsection{Data Processing Pipeline}

\textbf{Task instructions.}
We first define the creation objective and use GPT-6-astra to expand it into a task instruction.
The instruction specifies what the agent should construct and how the result should satisfy the intended geometry and appearance.
It also makes spatial relationships explicit so that the instruction describes a coherent asset or environment.
For tasks with visual references, the text complements the images with construction requirements; text-only tasks express the target entirely through the instruction.

\textbf{Reference images.}
We prepare visual inputs from existing 3D assets and collected imagery.
When a source 3D model or scene is available, we render it in Blender to obtain the required reference views.
Photographs and game screenshots provide references for environments captured outside this rendering workflow.
Where panoramic imagery is used, we project it into rectilinear views that show different directions from the same capture position.
The resulting images are paired with the task instruction to define the visual reconstruction target.

\textbf{Initial scenes for editing.}
We construct each editing task from a complete source scene.
The agent receives an initial Blender project that differs from the source scene in designated target components, a rendered reference of the target scene, and an instruction describing the requested edit.
The instruction identifies the content to modify and the surrounding content to preserve.
The source scene serves as the evaluation target for both the edited components and the preserved content.

\textbf{Environment and interaction specifications.}
We use GPT-6-astra to develop open-world instructions from the intended environment and activity.
Each instruction defines how functional regions connect and what the player must be able to accomplish within them.
When images are supplied, it distinguishes visible reconstruction targets from additional spaces and mechanisms that must be constructed to support the activity.
An accompanying interaction contract specifies the entities and their initial states, then defines the available operations and their physical constraints.
Action prerequisites and persistent state changes connect individual interactions into an activity that can be completed across the environment.

\subsection{Evaluation}
\label{sec:evaluation}

Frontier3DBench evaluates the appearance and structure of objects, the composition and editing of scenes, and the visual and functional coherence of interactive worlds.
Table~\ref{tab:score_components} defines the score components used across these levels.
Task-specific component weights are specified in Appendix~\ref{app:score_weights}.

\input{Tables/TabScoreComponents}

\subsubsection{Object-Level Scoring}
\label{sec:l1_evaluation}

L1 evaluates appearance fidelity, structural consistency, and detail fidelity.
Text-driven generation uses visual fidelity $V$, requirement satisfaction $R$, and detail fidelity $D$.
Reference reconstruction replaces requirement satisfaction $R$ with image similarity $I$ and object localization $L$, grounding appearance and placement in the supplied views.
The evaluator combines inspection of the delivered Blender scene with rendered evidence judged by GPT-6-astra at medium reasoning effort, with generator identity hidden.
Detail fidelity is measured on both the native scene and the exported asset.

\subsubsection{Scene-Level Scoring}
\label{sec:l2_evaluation}

L2 evaluates scene appearance, spatial organization, and completion of the requested construction or edit.
Brief2Scene uses the text-generation score (Eq.~\ref{eq:app_text_quality}), and Ref2Scene and Ref2Scene-RealPhoto use the reference score (Eq.~\ref{eq:app_reference_quality}), with native scene measurements and main and alternate rendered views.
Scene editing additionally assesses target modification $T$ and preservation $P$, comparing the candidate with the initial scene and target reference to determine whether the requested change is achieved while protected content is retained.

\subsubsection{Open-World-Level Scoring}
\label{sec:l3_evaluation}

L3 evaluates visual quality $A$, task fidelity $F$, spatial layout $G$, and interaction correctness $H$.
The submitted Unreal Engine project is rebuilt with an independent observer that captures scene views and records geometry, action effects, character movement, and state changes.
Local tests inspect individual mechanisms, while continuous workflows assess whether the environment supports the complete activity without teleportation or reset.

\subsubsection{Overall Score}
\label{sec:score_aggregation}

Task scores are reported on a 0--100 scale and averaged within each level, including measured zeros.
The overall score combines the three level means:
\begin{equation}
S_{\mathrm{overall}}=0.30S_{\mathrm{L1}}+0.30S_{\mathrm{L2}}+0.40S_{\mathrm{L3}}.
\label{eq:overall_score}
\end{equation}
The higher weight on open worlds emphasizes the integration of visual construction, interactive mechanisms, and complete activities.
Measurement procedures are detailed in Appendix~\ref{app:evaluation_details}.

%% file: Figures/FigBenchmarkComposition.tex
\begin{table}[htbp]
  \centering
  \caption{Frontier3DBench task families, counts, inputs, and requirements.}
  \label{tab:benchmark_composition}
  \begingroup
  \small
  \setlength{\tabcolsep}{4pt}
  \renewcommand{\arraystretch}{1.22}
  \begin{tabularx}{\linewidth}{@{}>{\raggedright\arraybackslash}p{0.195\linewidth}r>{\raggedright\arraybackslash}p{0.25\linewidth}>{\raggedright\arraybackslash}X@{}}
    \toprule
    \textbf{Task family} & \textbf{Tasks} & \textbf{Given input} & \textbf{Requirement} \\
    \midrule
    \rowcolor{black!5}
    \textbf{L1 · Objects} & \textbf{65} & \multicolumn{2}{l@{}}{\textit{Output: editable Blender assets}} \\
    Text2Code & 15 & Text instruction & Model the specified geometry and parts. \\
    \addlinespace[2pt]
    Img2Code & 25 & Multi-view images + instruction & Reconstruct an object across views. \\
    \addlinespace[2pt]
    Img2Code-SingleView & 25 & Single image + instruction & Reconstruct an object from one view. \\
    \addlinespace[5pt]
    \rowcolor{black!5}
    \textbf{L2 · Scenes} & \textbf{65} & \multicolumn{2}{l@{}}{\textit{Output: editable Blender scenes}} \\
    Brief2Scene & 20 & Text instruction & Compose objects, layout, and lighting. \\
    \addlinespace[2pt]
    Ref2Scene & 10 & Rendered image + instruction & Reconstruct the scene and appearance. \\
    \addlinespace[2pt]
    Ref2Scene-RealPhoto & 25 & Real photograph + instruction & Recover spatial layout and contents. \\
    \addlinespace[2pt]
    SceneEdit & 10 & Initial scene + reference + instruction & Edit the target while preserving surroundings. \\
    \addlinespace[5pt]
    \rowcolor{black!5}
    \textbf{L3 · Open worlds} & \textbf{15} & \multicolumn{2}{l@{}}{\textit{Output: interactive Unreal Engine projects}} \\
    Text2Env & 5 & Text instruction + interaction contract & Build connected spaces with the specified interactions. \\
    \addlinespace[2pt]
    Img2Env & 10 & Images + instruction + interaction contract & Reconstruct visible surroundings and required interactions. \\
    \bottomrule
  \end{tabularx}
  \endgroup
\end{table}

%% file: Tables/TabScoreComponents.tex
\begin{table}[htbp]
\centering
\caption{Evaluation components and their measurement scope. Numerical weights are given in Appendix~\ref{app:score_weights}.}
\label{tab:score_components}
\begingroup
\small
\renewcommand{\arraystretch}{1.12}
\setlength{\tabcolsep}{5pt}
\begin{tabularx}{\linewidth}{@{}ll>{\raggedright\arraybackslash}X@{}}
\toprule
Level & Component & What it measures \\
\midrule
L1/L2 & $V$: visual fidelity & Fidelity to the instruction or reference and aesthetic quality, judged from rendered views using normalized 0--4 grades. \\
 & $R$: requirements & Satisfaction of task-specific dimensions, part counts, geometric features, and spatial relations, checked in native scenes or, for Brief2Scene, with an evidence checklist. \\
 & $D$: detail fidelity & Coverage and implementation accuracy of required details; the weakest measured execution or export variant determines the score. \\
 & $I$: image similarity & Agreement with the input reference in image structure, edges, and color or tone over the prescribed image regions. \\
 & $L$: object localization & Projected object placement, size, and count, using matched reference and candidate boxes with penalties for missing or extra instances. \\
 & $T$: target modification & Whether the requested edit is achieved in the designated region, within the permitted geometry, material, or lighting changes. \\
 & $P$: preservation & Whether protected content agrees with the initial scene outside the permitted edit scope. \\
\midrule
L3 & $A$: visual quality & Shape and construction, materials, lighting, composition, detail, and consistency with the requested visual style. \\
 & $F$: task fidelity & Coverage of the required environment and visible identity of its functional areas and mechanisms. \\
 & $G$: spatial layout & Geometric support connectivity, static penetration, traversability through actual character movement, and layout semantics. \\
 & $H$: interaction correctness & Observable operation of local mechanisms, clearance of moving parts, and completion of required states and continuous workflows. \\
\bottomrule
\end{tabularx}
\endgroup
\end{table}

%% file: Sections/3_experiments.tex
\section{Experiments}
\label{sec:experiments}

\subsection{Experimental Setting}
\label{sec:experimental_setting}

We evaluate 11 frontier models: GPT-6.1-sol, GPT-6-astra, GPT-6-luna, Claude Opus 5.5, Kimi-K3, Kimi-K3.1-Pro, MiniMax-M3.1-Flash, DeepSeek-v4.1-Flash, GLM-5.3-Flash, Doubao-Seed-Evolving, and Qwen3.8-Max. Each model runs in its developer's agent harness: Codex for the GPT models, Claude Code for Claude Opus 5.5, Kimi CLI for the Kimi models, DeepSeek Harness (DSH) for DeepSeek-v4.1-Flash, and ZCode for GLM-5.3-Flash.
MiniMax-M3.1-Flash, Qwen3.8-Max, and Doubao-Seed-Evolving, which do not provide their own harness, run in Claude Code.
Each configuration uses the same harness for all three levels.
In the text, we abbreviate GPT-6.1-sol, GPT-6-astra, GPT-6-luna, and Claude Opus 5.5 as Sol, Astra, Luna, and Opus, and refer to other models by family name where unambiguous.

All models use their highest available reasoning effort: xhigh for Qwen and max for the others. Each agent decides when its work is ready for submission. If a submission fails the public execution and delivery checks, the agent can retry with the execution errors but without score feedback, within a fixed time budget (Appendix~\ref{app:execution_protocols}).

\subsection{Overall Performance}
\label{sec:overall_performance}

\paragraph{The best overall configuration combines strong object and scene performance, while category leaders vary.}
GPT-6.1-sol with Codex achieves the highest overall score of 66.02, with level means of 71.21 on objects, 70.35 on scenes, and 58.89 on open worlds.
Its lead over the runner-up is 2.17 points, with scene construction contributing the largest weighted advantage.
Across the 9 task categories, however, the highest scores are shared by three configurations: Sol leads six, Opus leads Text2Code, and Astra leads Text2Env and Img2Env.
No single configuration leads every category.
\input{Tables/TabOverall}

\paragraph{Reference reconstruction and interactive world construction remain challenging across configurations.}
Every configuration scores higher on text-to-object generation than on either object-reconstruction category, and higher on brief-driven scene generation than on either scene-reconstruction category.
Text-to-object scores range from 75.93 to 84.05, whereas even the best multi-view and single-view reconstruction scores reach only 69.61 and 66.26, respectively.
Every configuration also has a lower mean on L3 than on L1 or L2, with the highest open-world mean reaching 61.28.
Matching a specified reference and constructing an interactive world thus remain harder than generating plausible content.

\subsection{Model Capabilities across Tasks}
\label{sec:task_specialization}

\input{Figures/FigCategoryRadarFigure}

Figure~\ref{fig:category_radar} shows each model's profile across the 9 task categories relative to the cross-model mean.
We examine these profiles alongside Table~\ref{tab:overall_performance} to identify strengths that the overall ranking can obscure.

\paragraph{Task leaders vary across reconstruction, scenes, and worlds.}
Sol leads both object-reconstruction categories and all four scene categories, while Astra leads both open-world categories with the highest L3 mean of 61.28.
Opus instead achieves the highest text-to-object score of 84.05.
These complementary strengths make task-specific profiles useful for model selection.

\paragraph{Text-driven generation and reference reconstruction show distinct strengths.}
Kimi-K3.1-Pro and DeepSeek rank second and third in both text-to-object and brief-driven scene generation, but trail the leading models on reference reconstruction.
Qwen makes this gap especially clear: its text-to-object score of 81.98 nearly matches Sol's 82.10, whereas its multi-view and single-view reconstruction scores fall to 43.97 and 37.34.
Strong text-driven generation therefore does not imply accurate reconstruction of specified geometry and appearance.

\paragraph{Scene editing reveals strengths beyond the top overall ranks.}
MiniMax ranks second in editing at 78.70 despite placing sixth overall, ahead of both Astra and Opus.
Kimi-K3 and GLM also exceed the cross-model editing mean, scoring 73.22 and 72.91.
Their ability to modify existing scenes stands out relative to their weaker reference-reconstruction and open-world results.

\subsection{Model Performance across Capability Dimensions}
\label{sec:rubric_analysis}

\begin{figure}[htbp]
  \centering
  \includegraphics[width=\linewidth]{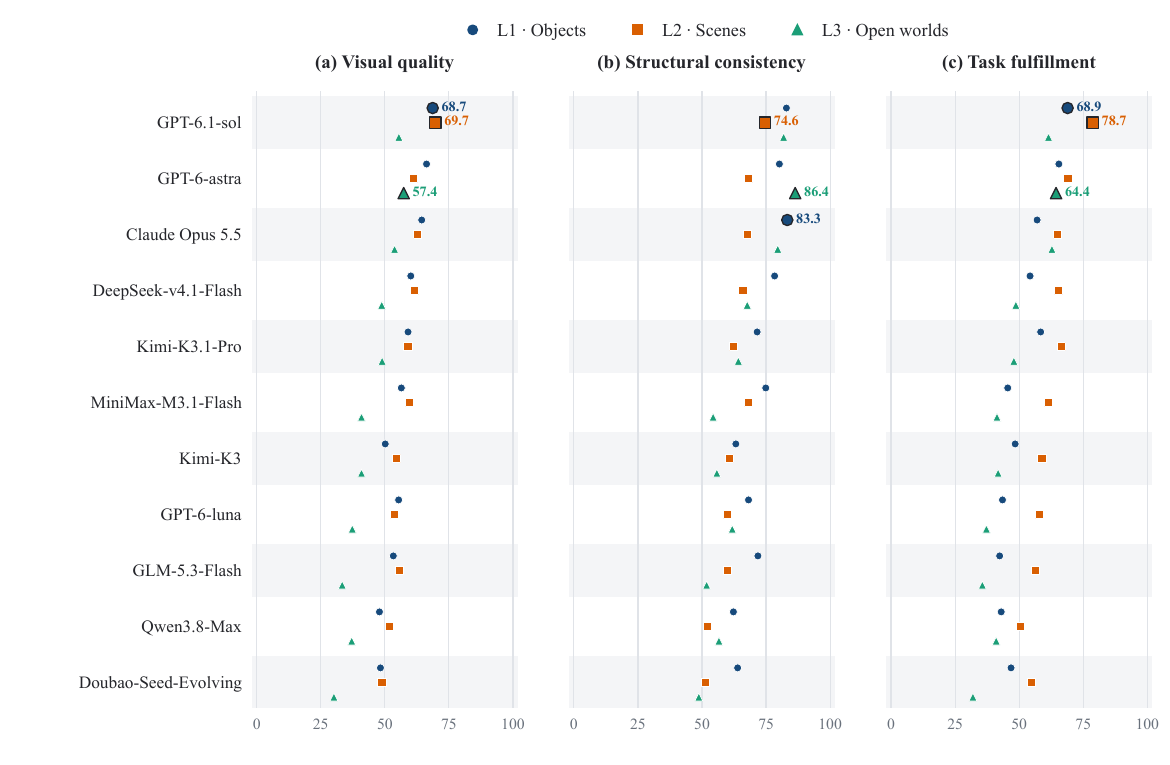}
  \caption{Capability profiles for visual quality, structural and spatial consistency, and task fulfillment (0--100). Markers show dimension means over 65 L1, 65 L2, and 15 L3 tasks; outlines and labels mark level-wise maxima. Models follow overall-score order.}
  \label{fig:rubric_profiles}
\end{figure}

We group the score components in Table~\ref{tab:score_components} into three capability dimensions.
Visual quality $S_{\mathrm{vis}}$ covers $V$ and $I$ in L1/L2 and $A$ in L3; structural and spatial consistency $S_{\mathrm{str}}$ covers $R$ and $L$ in L1/L2 and $G$ in L3; task fulfillment $S_{\mathrm{task}}$ covers $D$, $T$, and $P$ in L1/L2 and $F$ and $H$ in L3.
For each task, a dimension score is the weighted average of the components that the task uses, with weights given by their coefficients in Eqs.~\ref{eq:app_text_quality}--\ref{eq:app_world_quality} and normalized to sum to one within the dimension.
We then average each dimension within L1, L2, and L3.
Figure~\ref{fig:rubric_profiles} presents the resulting capability profiles on a 0--100 scale.

\paragraph{Visual leadership shifts from assets to worlds.}
Sol has the highest mean visual quality on objects and scenes, at 68.68 and 69.70, respectively.
Astra leads on open worlds at 57.37, followed by Sol at 55.49 and Opus at 53.80.
Object and scene appearance emphasize the requested form and reference agreement, while world appearance also spans lighting, composition, and style across a larger environment.

\paragraph{Structural consistency reveals a different ordering.}
Opus has the highest L1 structural score at 83.27, slightly above Sol at 82.93, despite trailing Sol in visual quality.
Sol leads L2 structure at 74.57, while Astra leads L3 layout at 86.37.
Across models, L1 structure averages 72.77 compared with 57.28 for visual quality and 52.15 for task fulfillment.
Coarse placement and structural requirements are thus easier to satisfy than complete appearance and fine detail.

\paragraph{Task fulfillment distinguishes detailed assets from usable worlds.}
Sol leads task fulfillment on objects and scenes at 68.90 and 78.72, whereas Astra leads worlds at 64.40, followed by Opus at 62.78 and Sol at 61.44.
This dimension measures detail completion in L1 and L2, adds target modification and preservation for scene editing, and combines environment coverage with interaction correctness in L3.
The world results show a gap between working parts and complete activities: across all 165 L3 submissions, the local-mechanism, motion-clearance, and state/workflow components of $H$ (Appendix~\ref{app:l3_measurements}) average 83.08, 20.82, and 30.26, respectively.

\subsection{Cost Efficiency Analysis}
\label{sec:cost_quality}

\begin{figure}[tbp]
  \centering
  \includegraphics[width=\linewidth]{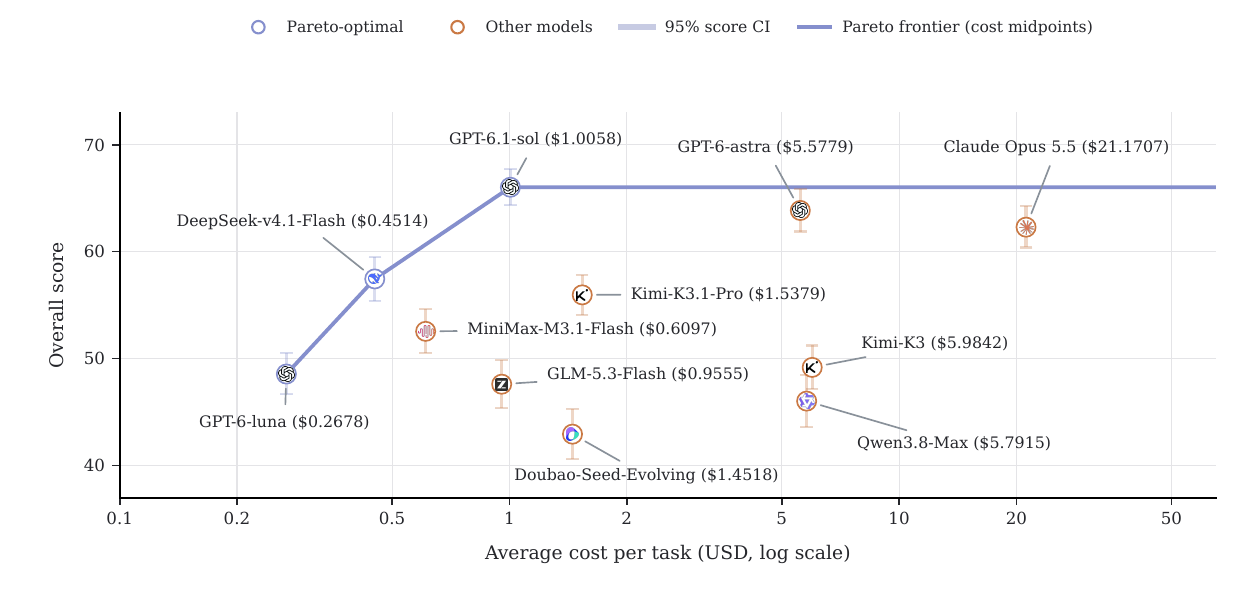}
  \caption{Overall score versus estimated average API cost per task across 145 tasks (log scale). Markers and the Pareto frontier use cost-range midpoints. Vertical bars show 95\% task-bootstrap intervals for the overall score.}
  \label{fig:cost_pareto}
\end{figure}

Figure~\ref{fig:cost_pareto} compares the overall score with estimated average API cost per task across all 145 tasks.
Costs cover recorded usage in each task's final successful generation attempt and weight every task equally, while the overall score uses the 30/30/40 level weights.
We use the midpoint of each exported cost range as its plotting coordinate; accounting and pricing assumptions are detailed in Appendix~\ref{app:cost_accounting}.

\paragraph{Luna, DeepSeek, and Sol define the score--cost frontier at the plotted estimates.}
Luna achieves 48.56 at approximately \$0.27 per task, DeepSeek achieves 57.45 at \$0.45, and Sol reaches the highest overall score of 66.02 at \$1.01.
Moving from DeepSeek to Sol gains 8.57 points at roughly 2.2 times the estimated cost.
DeepSeek's position reflects its low per-token prices and cache usage; because its usage records are partial, its estimated cost is a lower bound (Appendix~\ref{app:cost_accounting}).

\paragraph{Higher estimated spending does not consistently yield higher scores.}
Astra scores 63.85 at approximately \$5.58 per task, while Opus scores 62.29 at \$21.17.
Both cost more than Sol while scoring lower.
Kimi-K3.1-Pro scores 55.96 at \$1.54 per task, compared with DeepSeek's 57.45 at \$0.45.

\subsection{Qualitative Comparison}
\label{sec:qualitative_results}

\begin{figure}[htbp]
  \centering
  \includegraphics[width=\linewidth]{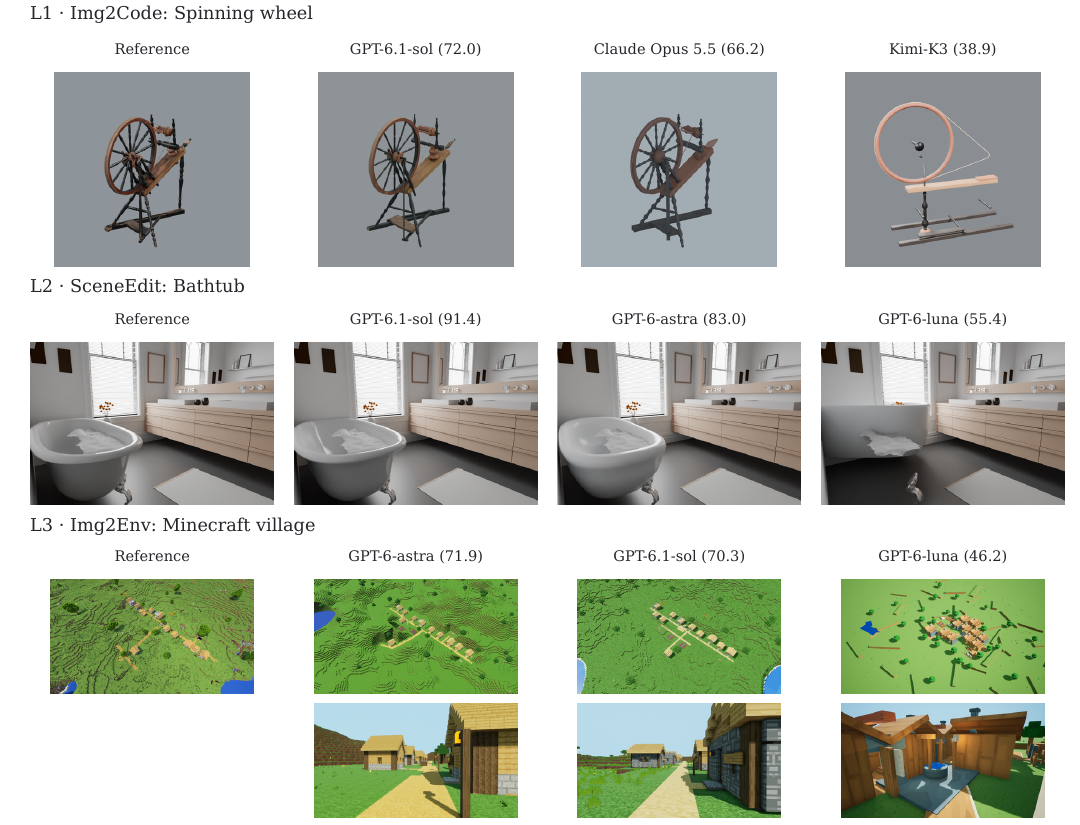}
  \caption{Object, scene, and interactive-world examples. Parentheses show task scores (0--100). World views are rendered from model-specific cameras. Minecraft reference: Akirby80.}
  \label{fig:case_studies}
\end{figure}

Figure~\ref{fig:case_studies} compares object reconstruction, scene editing, and reference-image environment creation.
For the spinning wheel, Sol (71.97) and Opus (66.21) recover the main mechanical assembly, whereas Kimi-K3 (38.87) reduces it to a ring-and-rod abstraction with missing components.
For the bathtub edit, Sol (91.43) and Astra (82.97) fit the tub to the retained supports, while Luna (55.42) places the shell apart from the feet and fails to accommodate the cloth.
For the Minecraft village, Astra (71.89) and Sol (70.29) reconstruct the village axis and continuous terraced landscape, though both simplify building variation; Luna (46.16) produces a dense cluster on largely flat terrain.
Across these cases, stronger outputs preserve component relationships and spatial context, while weaker outputs retain recognizable elements but lose important structure.

\subsection{Human--Score Agreement}
\label{sec:human_agreement}

To evaluate alignment with human visual judgments, we conduct a four-candidate ranking study with eight raters over 45 items from the four reference-conditioned L1/L2 task families (Img2Code, Img2Code-SingleView, Ref2Scene, and Ref2Scene-RealPhoto), yielding 360 judgments.

\paragraph{Automatic scores align with human fidelity rankings.}
Mean Kendall $\tau_b$ is 0.727, with 79.2\% top-one agreement and 44.7\% exact-order agreement (Table~\ref{tab:agreement-family}), indicating stronger agreement on the winner than on the complete ordering.
Human raters agree with one another at a mean $\tau_b$ of 0.679 across 1260 same-item pairs, so the automatic ranking agrees with individual raters at least as closely as raters agree with each other.
The 25 items spanning at least 30 score points reach mean $\tau_b=0.793$, compared with 0.644 for the other 20 items, so agreement increases with the score gap between candidates.

\input{Tables/TabHumanAgreement}
\FloatBarrier

%% file: Tables/TabOverall.tex
\begin{table}[tbp]
\centering
\caption{Performance on 145 tasks (0--100, $\uparrow$), ranked by Total, the overall score (Eq.~\ref{eq:overall_score}). T2C: Text2Code; I2C: Img2Code; SV: SingleView; B2S: Brief2Scene; R2S: Ref2Scene; RP: RealPhoto; SE: SceneEdit; T2E: Text2Env; I2E: Img2Env.}
\label{tab:overall_performance}
\begingroup
\fontsize{7}{8.4}\selectfont
\setlength{\tabcolsep}{1pt}
\renewcommand{\arraystretch}{1.23}
\newcommand{\mainhead}[1]{{\sffamily\bfseries #1}}
\newcommand{\taskhead}[1]{\mainhead{\shortstack{#1}}}
\begin{tabular*}{\linewidth}{@{\extracolsep{\fill}}lc@{\hspace{3pt}}|@{\hspace{3pt}}c@{\hspace{3pt}}|@{\hspace{3pt}}*{12}{c}@{}}
\toprule
\multirow{2}{*}{\mainhead{Model}} & \multirow{2}{*}{\mainhead{Harness}} & \multirow{2}{*}{\mainhead{Total}} & \multicolumn{4}{c}{\mainhead{L1: Objects}} & \multicolumn{5}{c}{\mainhead{L2: Scenes}} & \multicolumn{3}{c}{\mainhead{L3: Open worlds}} \\
\cmidrule(lr){4-7}\cmidrule(lr){8-12}\cmidrule(lr){13-15}
 & & & \taskhead{T2C} & \taskhead{I2C} & \taskhead{I2C\\(SV)} & \taskhead{Avg.} & \taskhead{B2S} & \taskhead{R2S} & \taskhead{R2S\\(RP)} & \taskhead{SE} & \taskhead{Avg.} & \taskhead{T2E} & \taskhead{I2E} & \taskhead{Avg.} \\
\midrule
\raisebox{-0.15ex}{\includegraphics[height=1.05em,page=1]{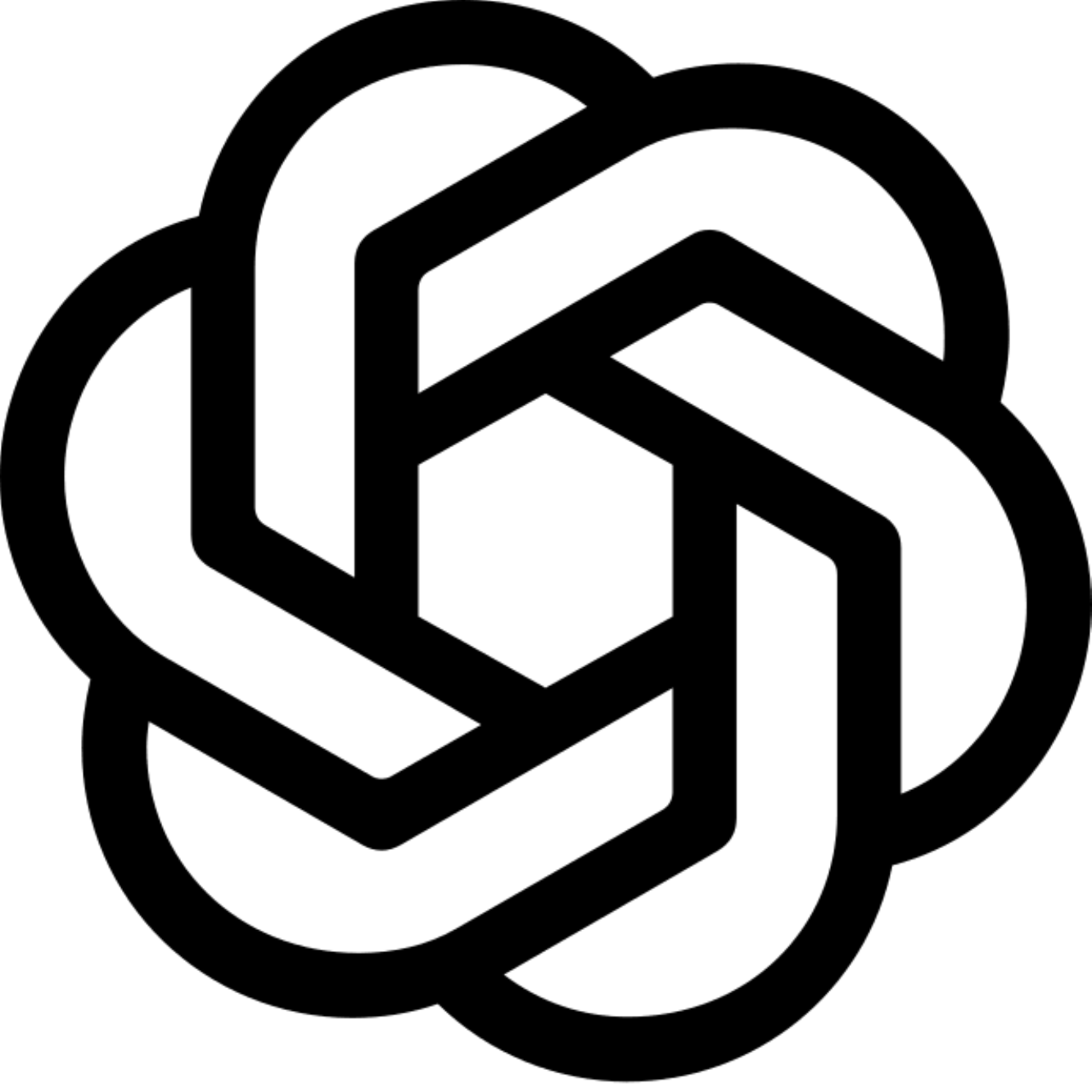}}\hspace{0.3em}GPT-6.1-sol & Codex & \textbf{66.02} & 82.10 & \textbf{69.61} & \textbf{66.26} & \textbf{71.21} & \textbf{79.87} & \textbf{60.95} & \textbf{60.86} & \textbf{84.44} & \textbf{70.35} & 61.31 & 57.68 & 58.89 \\
\raisebox{-0.15ex}{\includegraphics[height=1.05em,page=1]{Figures/ModelIcons.pdf}}\hspace{0.3em}GPT-6-astra & Codex & 63.85 & 79.54 & 68.44 & 62.28 & 68.63 & 74.71 & 48.03 & 53.91 & 74.11 & 62.51 & \textbf{64.07} & \textbf{59.88} & \textbf{61.28} \\
\raisebox{-0.15ex}{\includegraphics[height=1.05em,page=2]{Figures/ModelIcons.pdf}}\hspace{0.3em}Claude Opus 5.5 & Claude Code & 62.29 & \textbf{84.05} & 66.73 & 58.00 & 67.37 & 73.31 & 52.28 & 54.43 & 72.00 & 62.61 & 58.59 & 58.05 & 58.23 \\
\raisebox{-0.15ex}{\includegraphics[height=1.05em,page=3]{Figures/ModelIcons.pdf}}\hspace{0.3em}DeepSeek-v4.1-Flash & DSH & 57.45 & 83.17 & 57.49 & 56.92 & 63.20 & 76.36 & 42.44 & 52.73 & 76.00 & 62.00 & 49.91 & 49.63 & 49.72 \\
\raisebox{-0.15ex}{\includegraphics[height=1.05em,page=4]{Figures/ModelIcons.pdf}}\hspace{0.3em}Kimi-K3.1-Pro & Kimi CLI & 55.96 & 83.34 & 56.25 & 53.23 & 61.34 & 77.04 & 45.06 & 48.16 & 66.65 & 59.42 & 52.06 & 47.99 & 49.34 \\
\raisebox{-0.15ex}{\includegraphics[height=1.05em,page=5]{Figures/ModelIcons.pdf}}\hspace{0.3em}MiniMax-M3.1-Flash & Claude Code & 52.55 & 80.43 & 54.83 & 50.26 & 58.98 & 71.31 & 46.30 & 50.38 & 78.70 & 60.55 & 43.55 & 40.84 & 41.74 \\
\raisebox{-0.15ex}{\includegraphics[height=1.05em,page=4]{Figures/ModelIcons.pdf}}\hspace{0.3em}Kimi-K3 & Kimi CLI & 49.18 & 76.54 & 48.31 & 42.13 & 52.45 & 68.79 & 37.38 & 45.11 & 73.22 & 55.53 & 45.69 & 40.10 & 41.96 \\
\raisebox{-0.15ex}{\includegraphics[height=1.05em,page=1]{Figures/ModelIcons.pdf}}\hspace{0.3em}GPT-6-luna & Codex & 48.56 & 75.93 & 50.93 & 50.16 & 56.40 & 69.71 & 36.96 & 44.38 & 64.56 & 54.14 & 40.56 & 37.47 & 38.50 \\
\raisebox{-0.15ex}{\includegraphics[height=1.05em,page=6]{Figures/ModelIcons.pdf}}\hspace{0.3em}GLM-5.3-Flash & ZCode & 47.61 & 80.62 & 48.97 & 48.29 & 56.01 & 68.56 & 41.96 & 44.60 & 72.91 & 55.92 & 37.87 & 33.68 & 35.08 \\
\raisebox{-0.15ex}{\includegraphics[height=1.05em,page=7]{Figures/ModelIcons.pdf}}\hspace{0.3em}Qwen3.8-Max & Claude Code & 46.03 & 81.98 & 43.97 & 37.34 & 50.19 & 69.56 & 32.42 & 39.65 & 58.49 & 50.64 & 43.83 & 37.25 & 39.44 \\
\raisebox{-0.15ex}{\includegraphics[height=1.05em,page=8]{Figures/ModelIcons.pdf}}\hspace{0.3em}Doubao-Seed-Evolving & Claude Code & 42.94 & 78.60 & 43.71 & 42.36 & 51.24 & 66.52 & 33.39 & 37.63 & 61.74 & 49.58 & 28.71 & 33.23 & 31.72 \\
\bottomrule
\end{tabular*}
\endgroup
\end{table}

%% file: Figures/FigCategoryRadarFigure.tex
\begin{figure}[tbp]
  \centering
  \includegraphics[width=\linewidth]{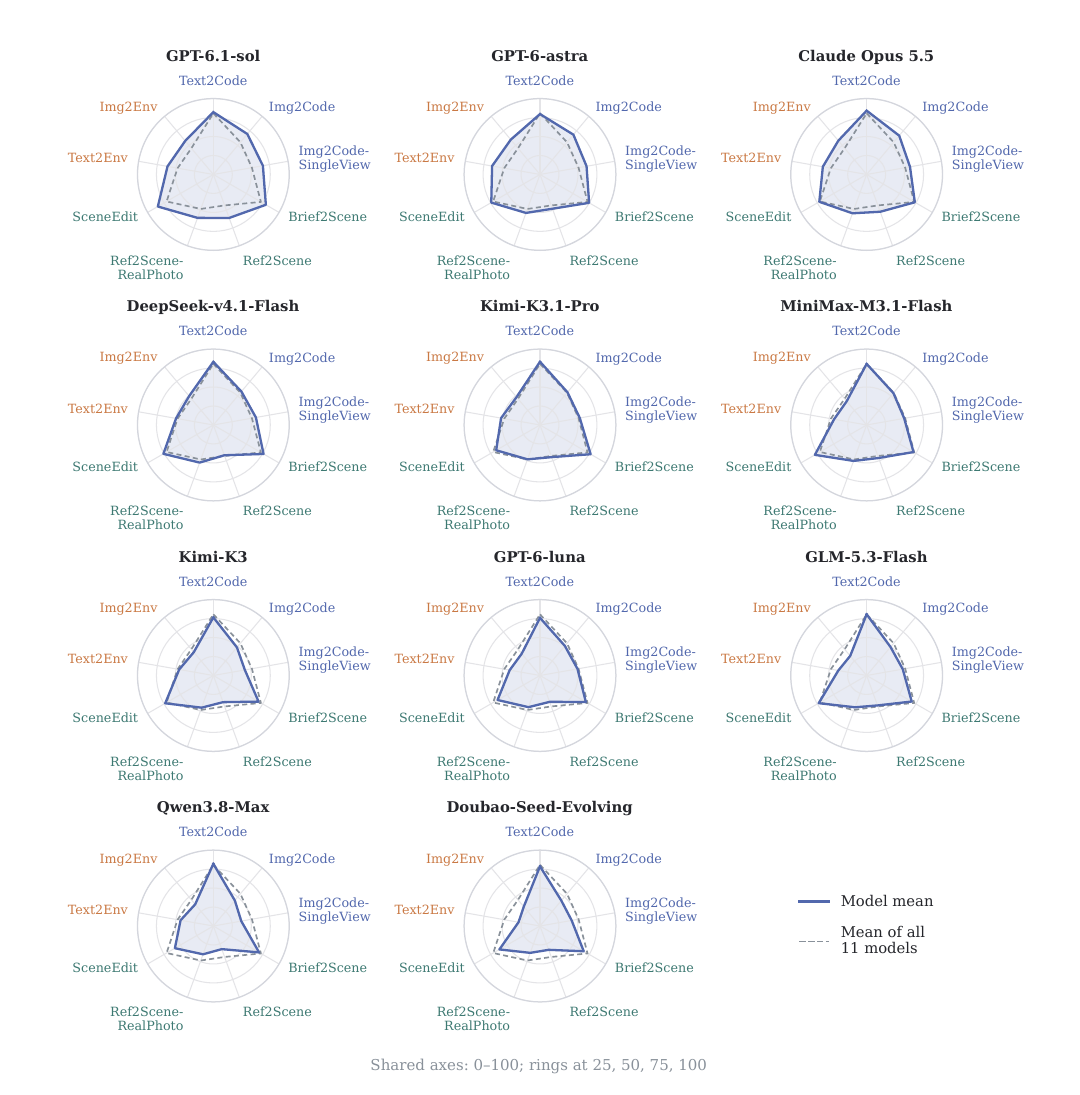}
  \caption{Task-category profiles in overall-score order. Solid contours show each model's task means; dashed contours show the mean over all 11 configurations.}
  \label{fig:category_radar}
\end{figure}

%% file: Tables/TabHumanAgreement.tex
\begin{table}[htbp]
\centering
\caption{Agreement between human rankings and automatic task scores on four reference-conditioned task families (eight raters, 45 items, 360 rankings).}
\label{tab:agreement-family}
\begingroup
\small
\setlength{\tabcolsep}{3.5pt}
\renewcommand{\arraystretch}{1.12}
\begin{tabular*}{\linewidth}{@{\extracolsep{\fill}}lrrrrr@{}}
\toprule
Family & Items & Judgments & Mean $\tau_b$ & Top-one & Exact order \\
\midrule
Img2Code & 14 & 112 & 0.753 & 77.7\% & 43.8\% \\
Img2Code-SingleView & 16 & 128 & 0.659 & 78.1\% & 39.1\% \\
Ref2Scene & 8 & 64 & 0.781 & 84.4\% & 50.0\% \\
Ref2Scene-RealPhoto & 7 & 56 & 0.768 & 78.6\% & 53.6\% \\
\midrule
All families & 45 & 360 & 0.727 & 79.2\% & 44.7\% \\
\bottomrule
\end{tabular*}

\endgroup
\end{table}

%% file: Sections/4_relatedworks.tex
\section{Related Work}
\label{sec:related_work}

\paragraph{Code-based 3D creation benchmarks.}
CAD benchmarks evaluate geometry, parametric structure, and engineering constraints from text and visual specifications~\cite{benchcad2026,benchcad22026,p3dbench2026,realcad2026,industrialcad2026}.
3DCodeBench and 3DHarnessBench assess procedural asset generation and reconstruction through Blender code~\cite{code3dbench2026,harness3d2026}, and GeoCodeBench~\cite{li2026benchmarking} benchmarks research-level coding for 3D geometric computer vision.
BlenderGym, VIGA, and MUSE study visual reconstruction and instruction-guided authoring or editing~\cite{blendergym2025,viga2026,authorbench2026}, while VoxelCodeBench and SceneActBench probe spatial composition and scene operations~\cite{voxelcode2026,sceneact2026}.
Code4Scene evaluates Unreal Engine scene construction and editing through artifact, static-physics, and auxiliary navigation checks~\cite{code4scene2026}.
BlenderWorld spans object and environment creation, reconstruction, modification, and animation, combining rendered evidence with editable-project inspection~\cite{blenderworld2026}.

\paragraph{Agents for interactive 3D creation.}
Recent systems connect object construction, scene organization, and physical interaction.
NeoWorld-3 reconstructs articulated objects and compositional scenes from visual observations through executable programs and physics validation~\cite{neoworld32026}.
SceneCode couples room planning with part-wise object code generation to produce editable indoor scenes with articulated assets~\cite{scenecode2026}.
EmbodiedGen V2 integrates asset generation, multi-room environments, task-driven worlds, and agent-based editing for embodied simulation~\cite{embodiedgenv22026}.
Code2Worlds-2 extends object and scene generation with physical compilation and simulator feedback, while Code2Games connects Blender world construction to UE5 gameplay through scene--gameplay correspondence and execution-guided repair~\cite{code2worlds22026,code2games2026}.
These systems show that coding agents can construct content across scales, which motivates evaluating each scale with dedicated tasks.

\paragraph{Dynamics and world evaluation.}
Video-conditioned benchmarks assess programmatic reconstruction of appearance, camera motion, geometry, and dynamics~\cite{bvb2026,cinerecon2026,code4dbench2026,physcodegym2026}; text-conditioned benchmarks assess generated simulations~\cite{physcodebench2026,simworlds2026}.
ExeWorldBench progresses from individual mechanisms to composed mechanical systems and goal-driven executable worlds, organizing evaluation around mechanical and causal dependencies~\cite{exeworld2026}.
Video2World and ScIntBench assess reconstruction from embodied demonstrations and physical interactions in generated scenes, respectively~\cite{video2world2026,scintbench2026}.
WorldClaw and VibeWorlding study open-world authoring~\cite{worldclaw2026,vibeworlding2026}, and VibeWorlding's VWE-Bench uses rubric-based verification of physical feasibility and intent fulfillment.
Browser-world evaluations combine visual assessment, code inspection, or runtime behavioral checks~\cite{aa3d2026,worldbench2026,worldcoder2026}.
Game benchmarks assess integrated assets, scenes, and mechanics~\cite{autoue2026,gameasg2026,code2games2026,cegame2026}, and CEGame-Bench additionally checks completion and saved state across engines.
CraftBench-UE and GenXR-Bench evaluate executable engine and XR features~\cite{craftbenchue2026,genxr2026}, while WorldScore evaluates sequential world generation across 3D, 4D, and video systems~\cite{worldscore2025}.

Frontier3DBench organizes coding-agent evaluation around dedicated object, scene, and interactive-open-world tasks within a shared benchmark.
Its hierarchy follows the unit being constructed and the corresponding correctness requirements: object geometry and parts, scene layout and composition, and connected world regions with stateful interaction sequences.
All three levels cover generation and reconstruction; L2 additionally covers editing, and L3 adds interactive activities with persistent state.
This design allows comparison across levels while each level keeps its own visual, structural, and behavioral checks.

%% file: Sections/5_conclusion.tex
\section{Conclusion}
\label{sec:conclusion}

We introduced Frontier3DBench to measure how far coding agents have come in 3D creation.
Its 145 tasks across 9 categories unify object, scene, and interactive-world construction, while evaluation of executable artifacts jointly assesses appearance, structure, and behavior.
Across 11 model--harness configurations, leadership shifts from GPT-6.1-sol on objects and scenes to GPT-6-astra on interactive worlds.
However, all configurations score lower on reference reconstruction than on text-driven generation in object and scene tasks, and strong local interactions remain insufficient for completing coordinated activities.
The central challenges are therefore matching specified geometry and layouts, and coordinating moving parts, actions, and persistent state.
Frontier3DBench makes these gaps measurable across construction levels, supporting progress toward coding agents that create faithful, coherent, and functional 3D content.

%% file: Sections/appendix_score_weights.tex
\section{Scoring Details}
\label{app:score_weights}

This appendix specifies the numerical weights and measurement procedures for the evaluation components defined in Section~\ref{sec:evaluation} and Table~\ref{tab:score_components}.
All components lie in $[0,1]$, and the task score is $s=100Q$.

\subsection{Task-Specific Quality Scores}

\paragraph{Text-driven object and scene generation.}
For \textit{Text2Code} and \textit{Brief2Scene}, visual fidelity $V$, requirement satisfaction $R$, and detail fidelity $D$ give
\begin{equation}
Q_{\mathrm{text}}=0.60V+0.30R+0.10D.
\label{eq:app_text_quality}
\end{equation}
$V$ measures fidelity to the instruction and aesthetics, $R$ measures task-specific geometric and spatial requirements, and $D$ measures the coverage and accuracy of fine details.

\paragraph{Reference-based object and scene reconstruction.}
For \textit{Img2Code}, \textit{Img2Code-SingleView}, \textit{Ref2Scene}, and \textit{Ref2Scene-RealPhoto}, the score is
\begin{equation}
Q_{\mathrm{ref}}=0.20V+0.50I+0.20L+0.10D.
\label{eq:app_reference_quality}
\end{equation}
Here $V$ measures visual fidelity to the input reference, $I$ measures image similarity, $L$ measures projected object placement and count, and $D$ measures reference-detail fidelity.

\paragraph{Scene editing.}
For \textit{SceneEdit}, the score is
\begin{equation}
Q_{\mathrm{edit}}=0.20V+0.20R+0.10D+0.30T+0.20P.
\label{eq:app_edit_quality}
\end{equation}
The additional terms $T$ and $P$ measure achievement of the requested target modification and preservation of protected content, respectively.

\paragraph{Interactive open worlds.}
Both image- and text-conditioned L3 tasks use
\begin{equation}
Q_{\mathrm{world}}=0.60A+0.20F+0.05G+0.15H.
\label{eq:app_world_quality}
\end{equation}
$A$ measures visual quality, $F$ task fidelity, $G$ spatial layout, and $H$ interaction correctness.

\subsection{Component Measurements}
\label{app:evaluation_details}

Task-specific checks validate execution, delivery, and artifact integrity.
Confirmed failures receive zero; valid artifacts receive component-wise quality scores.

\subsubsection{Blender and Image Measurements}
\label{sec:rule_evaluation}

The evaluator executes each submitted program in Blender 5.2.1 for a baseline and three seed reruns, checking required files, scene loading, editable geometry, and dependencies.
Measurements combine native scene data with rendered evidence.
$R$ measures task-specific dimensions, parts, geometry, and spatial relations against declared tolerances; for Brief2Scene, these requirements are specified as an evidence checklist.
For editing, $T$ measures the requested change and $P$ measures preservation of protected content.

Reference similarity combines SSIM~\citep{wang2004ssim}, color, tone, and edges:
\begin{equation}
I_{\mathrm{global}}=0.45I_{\mathrm{SSIM}}+0.25I_{\mathrm{color}}+0.15I_{\mathrm{tone}}+0.15I_{\mathrm{edge}}.
\label{eq:global_image}
\end{equation}
SSIM averages the original and two blurred scales; color and tone use a $4\times4$ grid, and edges use symmetric distance.
Region-level similarity combines SSIM, edge F1, and color with weights $0.50$, $0.30$, and $0.20$, respectively, attenuated by edge-density and contrast agreement.
$I$ combines region-level and global similarity with weights $0.90$ and $0.10$; tasks with static reference regions use the region term alone.
Localization $L$ matches fixed reference boxes to visually grounded candidate boxes using overlap, center proximity, and size with weights $0.50$, $0.30$, and $0.20$, respectively, penalizing missing and extra instances.
It measures projected placement and count.

\subsubsection{Visual Judgments and Detail}
\label{sec:vlm_evaluation}

GPT-6-astra with medium reasoning effort judges predefined criteria from the task instruction and rendered evidence views, together with the input references or, for SceneEdit, the initial and target views.
Generator identity is hidden from the judge at all levels.
Visual fidelity is $V=\sum_j a_jg_j/4$, where $\sum_j a_j=1$ and $g_j\in\{0,1,2,3,4\}$ denotes absent, weak, partial, mostly correct, or fully demonstrated fulfillment.
Grades require evidence-linked observations.
Each submission is scored with its first valid judgment.

For feature coverage $c_{jv}\in[0,1]$, implementation deviation $e_{jv}\in[0,1]$, and measured variant set $\mathcal V$, detail fidelity is
\begin{equation}
D=\min_{v\in\mathcal V}\sum_j b_jc_{jv}(1-e_{jv}),
\qquad \sum_j b_j=1.
\label{eq:detail_fidelity}
\end{equation}
We set $e_{jv}=0$ whenever $c_{jv}=0$.
The variant set $\mathcal V$ contains eight L1 variants (the baseline and three seed runs, each as a native scene and a GLB export) or four native L2 variants.

\subsubsection{Open-World Measurements}
\label{app:l3_measurements}

The evaluator rebuilds the submitted Unreal Engine project with an independent observer in a separate workspace and measures rendered appearance, geometry, motion, character trajectories, and inventory changes.
Both image- and text-conditioned tasks are scored with Eq.~\ref{eq:app_world_quality}.
The visual judge receives the instruction, references, the task's fixed reference-feature list, and independently rendered views, with generator identity hidden.
$A$ averages six criteria: shape and construction, materials, lighting, composition, detail, and style.
$F$ averages scene coverage and the visible identity of functional areas and mechanisms.
These criteria are scored directly with evidence anchors whose upper bounds are $0$, $0.34$, $0.64$, $0.84$, and $1$ for absent through exceptional realizations.

$G$ combines support connectivity, static penetration, traversability, and layout semantics with weights $0.25$, $0.25$, $0.30$, and $0.20$, respectively.
Support measures geometric connectivity to inferred ground; traversability uses actual character movement.
$H$ combines local mechanisms, motion clearance, and state/workflow checks with weights $0.40$, $0.25$, and $0.35$, respectively.
Local tests weight entity presence, observable action effects, and invalid-reach rejection by $0.20$, $0.50$, and $0.30$, respectively, averaging within and across mechanism roles.
The Minecraft task uses task-specific probes with the same weights.
Motion-clearance checks test sampled moving parts for intersections.
State/workflow equally weights satisfied end-state predicates and complete-route success.
Local probes can place test fixtures, whereas continuous workflows prohibit teleportation and resets.

%% file: Sections/appendix_experiments.tex
\section{Experimental Protocols}
\label{app:experimental_details}

\subsection{Tasks and Runtime Configurations}
\label{app:execution_protocols}

We evaluate 145 tasks: 65 object tasks (L1), 65 scene tasks (L2), and 15 interactive-world tasks (L3).
Each of the 11 model--harness configurations completes the full suite, and all 1,595 task scores are fully measured.
Models use their highest available reasoning effort, as listed in Section~\ref{sec:experimental_setting}.
Agents receive the English task instruction and public task assets; evaluator files and scores are not accessible during generation.
At all levels, general network access is disabled during generation and evaluation, and model requests use a restricted host gateway.
L1/L2 generation allows at most six attempts within a 32-hour budget per task; a new attempt starts after a timeout, a transport error, or a failed public execution or delivery check, and receives the execution errors without score feedback.
L3 agents work under a 24-hour time limit.
The released code uses longer identifiers for some task families and harnesses (e.g., \texttt{Image2Code}, \texttt{Reference2Scene}, \texttt{Image2Environment}, and \texttt{kimi-codecli}), which correspond one-to-one to the names used here.

\paragraph{Object and scene tasks.}
Generation runs in isolated Docker containers with two CPU cores and 4\,GiB of memory.
Submitted source bundles are executed in headless Blender 5.2.1.
L1 deliverables include an editable native scene and a GLB export; L2 deliverables include a native scene and main and alternate renders.
Rendering settings follow each task's specification.
Visual judgments use GPT-6-astra with medium reasoning effort under the scoring protocol in Appendix~\ref{app:evaluation_details}.

\paragraph{Interactive-world tasks.}
Agents develop Unreal Engine 5.8.3 projects in persistent Docker workspaces with eight CPU cores, 64\,GiB of memory, and an assigned GPU.
Public inputs include the instruction, applicable reference images, the interaction contract, integration interfaces, and read-only task-appropriate asset libraries.
The visual judge is GPT-6-astra with medium reasoning effort.

\subsection{Cost Accounting and Uncertainty}
\label{app:cost_accounting}

Figure~\ref{fig:cost_pareto} reports the estimated average cost of each task's final successful generation attempt.
For each task, recorded ordinary input, cache-read input, cache-write input, and output tokens are priced separately, then costs are averaged equally across the 145 tasks.
Cached tokens are included in input counts and are assigned their respective rates without double counting; reasoning tokens are included in output.
Prices follow the providers' Standard API pricing as of October 30, 2026.
CNY prices are converted at 6.6684 CNY per USD using the November 5, 2026 exchange rate; USD quotes require no conversion.
Qwen's cache-read price is set to 10\% of its ordinary input price.

Cost ranges reflect variation in cache-write lifetime, peak and off-peak pricing, and long-context tiers.
For aggregated input counts above a context threshold, the long-context multiplier is applied to the upper bound.
The figure plots range midpoints for model positions and the Pareto line and omits horizontal ranges.
Usage records are partial for DeepSeek, both Kimi configurations, and GLM, so their estimates are lower bounds.
The estimates cover generation API usage; judging, tool use, infrastructure, cache storage, and taxes are not included.

The vertical bars show 95\% task-bootstrap intervals for the overall score.
We jointly resample task identities across all 11 configurations within each level, drawing 65 L1, 65 L2, and 15 L3 tasks with replacement and retaining the 0.30/0.30/0.40 weights.
The intervals use 100,000 resamples with seed 20261105 to estimate task-sampling variation for the evaluated submissions.

\subsection{Human-Evaluation Protocol}
\label{app:human_protocol}

The 45 items comprise 14 multi-view objects, 16 single-view objects, eight rendered-reference scenes, and seven photographic-reference scenes.
Each item contains the highest- and lowest-scoring valid runs and the two nearest the score interval's trisection points, with adjacent options at least four points apart.
Eight raters rank every item by reference fidelity, producing 360 complete rankings.

Raters view the reference and four equally sized candidate images labelled A--D, with model identities and scores hidden.
Candidate labels are permuted independently for each rater and item.
Kendall's $\tau_b$, top-one agreement, and exact-order agreement compare human and score rankings of the same candidates; rater-to-rater agreement uses all $45\binom{8}{2}=1260$ same-item pairs.

%% file: Sections/appendix_visual_results.tex
\clearpage
\section{More Visual Results}
\label{app:visual_results}

\begin{figure}[!htbp]
  \centering
  \includegraphics[width=0.97\linewidth]{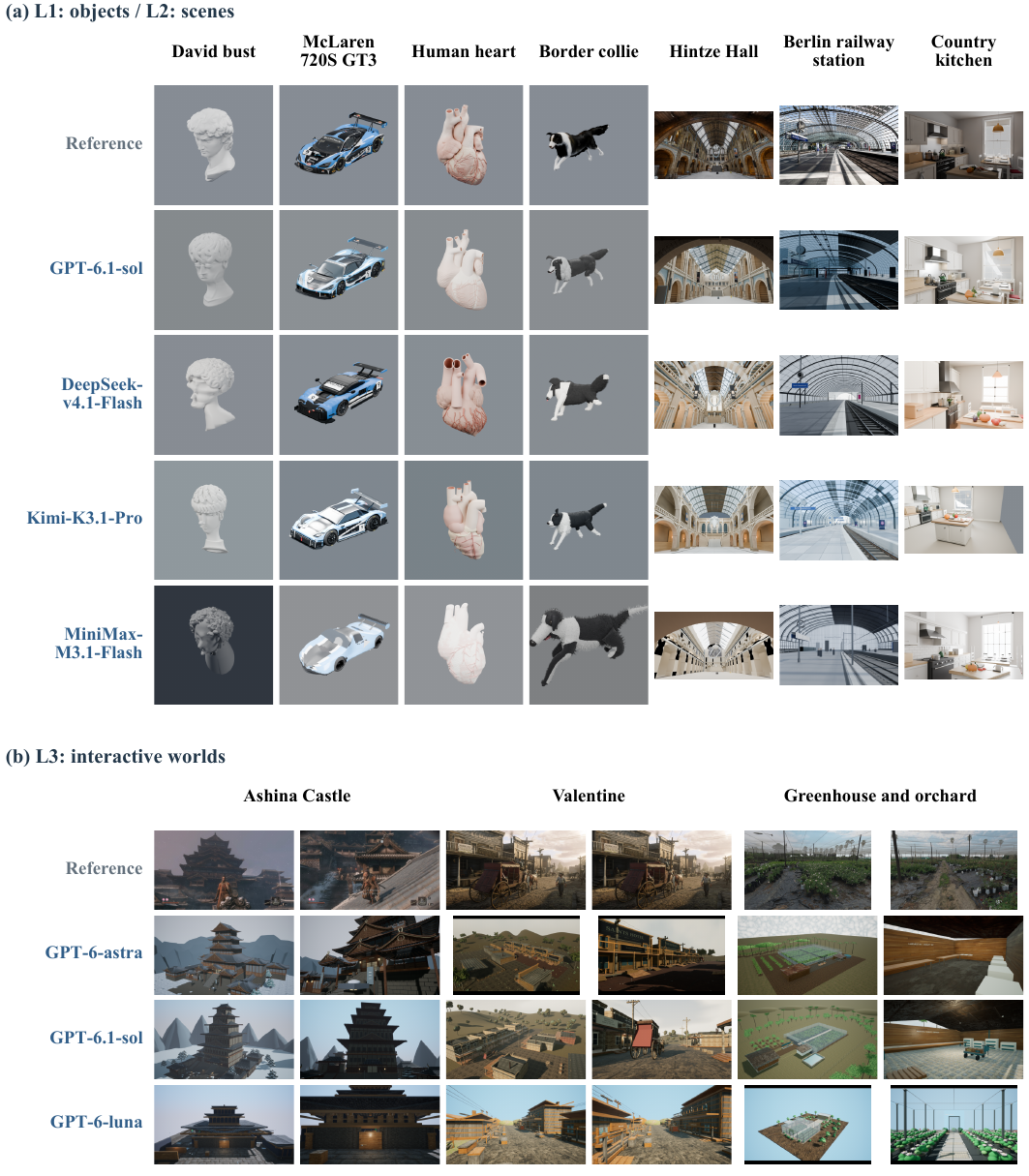}
  \caption{\textbf{Additional visual comparisons across objects, scenes, and interactive worlds.}
  Columns group cases; rows show references and model outputs.
  \textbf{(a)} Four object and three scene cases compare runs of GPT-6.1-sol, DeepSeek-v4.1-Flash, Kimi-K3.1-Pro, and MiniMax-M3.1-Flash.
  \textbf{(b)} Three world cases each show two submitted project previews from GPT-6-astra, GPT-6.1-sol, and GPT-6-luna.
  Cases are drawn from the 145-task suite.
  World views use model-specific cameras; Valentine uses the same reference for both views.
  World reference credits: FromSoftware/Activision, Rockstar Games, and Poly Haven (CC0).}
  \label{fig:appendix_visual_results}
\end{figure}
\FloatBarrier